\documentclass[11pt]{article}

\usepackage[final]{acl}

\usepackage{times}
\usepackage{latexsym}
\usepackage{tabularray}
\usepackage[T1]{fontenc}

\usepackage[utf8]{inputenc}

\usepackage{microtype}

\usepackage{inconsolata}

\usepackage{graphicx}
\usepackage{booktabs, graphicx, makecell, xcolor, pifont}

\newcommand{\yesmark}{%
  {\setlength{\fboxsep}{1pt}%
   \textcolor{green!50!black}{%
     \fbox{\makebox[0.8em]{\strut\ding{51}}}}}}
\newcommand{\nomark}{%
  {\setlength{\fboxsep}{1pt}%
   \textcolor{red!75!black}{%
     \fbox{\makebox[0.8em]{\strut\ding{55}}}}}}

\title{Joi-Bench-Task: Benchmarking Task Coordination in Full-duplex Voice Agents through Interactive Simulation}

\author{Xiangheng He\textsuperscript{\(\dagger\)}, Mingjun Zhao\textsuperscript{\(\dagger\)},  Zhehao Fan,  Zhengjia Li,  \\
\textbf{Xiaopeng Sun, Qingqing Xiao, Haolan Chen\textsuperscript{\(\ddagger\)}, Yu Xu} \\
\\
Platform and Content Group, Tencent \\ }

\begin{document}
\maketitle
\begingroup
\renewcommand{\thefootnote}{}
\footnotetext{\textsuperscript{\(\dagger\)} Co-first authorship. 
  
\textsuperscript{\(\ddagger\)} Corresponding author. haolanchen@tencent.com.
}
\endgroup

\begin{abstract}
Full-duplex voice agents enable natural, hands-free interaction, allowing users to refine requests, redirect ongoing work, and seek clarification while the agent is speaking. However, existing voice-agent benchmarks insufficiently isolate coordination between task execution and conversation. We introduce a benchmark that assesses this coordination by examining whether agents track what is being executed, what has been completed, what has been communicated, and what remains to be reported as the conversation evolves. We propose an interactive user simulator that introduces configurable interventions at designated timing in tool execution and spoken result delivery, supporting controllable and reproducible evaluation conditions. We evaluate five leading commercial voice agents along three dimensions: execution responsiveness, coordination under request changes, and coordination under interruptions. Agent evaluations reveal persistent coordination failures. Under request changes, the highest result completion rate is only 55\%, indicating difficulty for current voice agents adapting to new requests while preserving earlier obligations. After user interruption, four of the five agents exhibit average information omission rates of at least 23\%. These findings highlight task coordination as a central challenge for current voice agents in complex, evolving conversations.

\end{abstract}

\section{Introduction}
\begin{figure*}[t]
  \includegraphics[width=\linewidth]{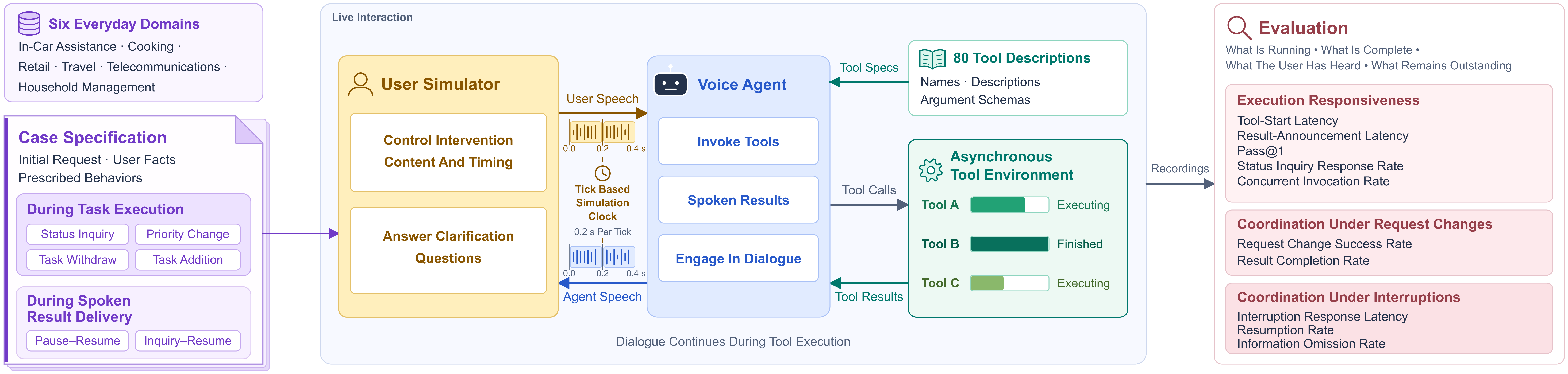} 
  \caption {The framework for Joi-Bench-Task.}
  \label{fig:benchmark_framework}
\end{figure*}

Large language model (LLM) agents combine reasoning, planning, and tool use to accomplish complex tasks \cite{DBLP:conf/iclr/YaoZYDSN023}. Full-duplex voice agents extend these capabilities to spoken interaction, processing speech while speaking and enabling users to interrupt at any moment \cite{defossez2024moshi}. This barge-in capability lets users refine requirements, redirect ongoing work, and seek clarification during long-horizon tasks, supporting more natural collaboration. Advances in speech interaction and tool use are moving voice agents beyond real-time dialogue prototypes toward practical assistants for complex everyday tasks \cite{DBLP:journals/corr/abs-2603-13686}.

Recent voice-agent benchmarks jointly assess conversational dynamics and task performance \cite{DBLP:journals/corr/abs-2603-13686,lin2026fullv3}, with GPT-Live-1 achieving 84.7\% completion on English tasks \cite{taubench_website}. Yet task success alone does not establish whether agents can coordinate ongoing work with evolving dialogue. While tools are still running, agents must promptly answer progress inquires and reconcile task additions, cancellations, or priority changes with ongoing operations. Likewise, when interrupted while reporting tool results, agents must accurately track what users have heard and resume after follow-ups or pauses without omitting key information. Evaluating these capabilities requires interventions timed to tool execution and speech playback, alongside assessment of how agents maintain execution and communication state across them. Existing evaluations insufficiently isolate this coordination.
 
Evaluating this coordination requires an interactive, controllable user simulator that acts as a proxy for a human user throughout the conversation. In realistic interactions, users may revise or expand their requests as tasks unfold, while also answering clarification questions from the agent. A simulator must therefore generate contextually appropriate responses and control when specific user behaviors occur. The timing of these behaviors is particularly important: adapting ongoing work should be tested during tool execution, whereas interrupting and resuming unfinished communication should be tested during result delivery. Conditioning user interventions on these time stages enables reproducible evaluation of how the agent remains responsive while tracking task and communication progress.
 
Guided by these evaluation requirements, we introduce a novel benchmark for assessing coordination in full-duplex voice agents during complex, long-horizon tasks, focusing on their ability to track what is being executed, what has been completed, what has been communicated, and what remains to be reported as the conversation evolves. Our contributions are threefold:

\begin{enumerate}
\item We propose a novel benchmark that evaluates how full-duplex voice agents coordinate ongoing task execution and spoken result delivery as user needs evolve through real-time interaction.
\item We develop a controllable user simulator that controls intervention timing according to both tool execution state and results announcement progress while remaining interactive during conversations, providing configurable and reproducible test conditions.
\item We systematically evaluate five leading commercial full-duplex voice agents  \footnote{API versions used: GPT Live 1.0 (September 10, 2026);
GPT Realtime 1.5 (February 23, 2026);
GPT Realtime 2 (May 7, 2026);
Grok Voice 3.0 (October 14, 2026);
Gemini 3.8 Live (September 15, 2026).}, characterizing their limitations and strengths regarding conversational dynamics and tool coordination when addressing evolving user needs.
\end{enumerate}

\section{Related Works}
\subsection{Voice Agent Benchmarks}
Earlier voice-agent benchmarks largely evaluated conversational dynamics \cite{lin2025fullv1,lin2026fullv15,lin2026fullv2,modi2026echochain,salimi2026ihbench} and agentic performance \cite{jain2025voiceagentbench} separately. More recent benchmarks integrate conversational dynamics with agentic task evaluation \cite{DBLP:journals/corr/abs-2603-13686,lin2026fullv3,bogavelli2026eva,meyer2026vamos,mathur2026apex}. 
$\tau$-Voice \cite{DBLP:journals/corr/abs-2603-13686} evaluates full-duplex task completion under realistic acoustic conditions. EVA-Bench \cite{bogavelli2026eva} measures task completion, faithfulness, speech fidelity, and conversational quality. VAmoS Bench \cite{meyer2026vamos} assesses action sequencing and consistency between spoken claims and tool outcomes. APEX-Voice \cite{mathur2026apex} evaluates professional workflows, artifacts, and adherence to corrections and authorization changes. 
However, these benchmarks have not systematically evaluated how voice agents track and coordinate ongoing tasks when user requirements change during tool execution or spoken result delivery is interrupted. Our benchmark addresses this gap by introducing controlled user interventions at both stages and evaluating task coordination at these stages. Table \ref{tab:benchmark-comparison} provides a detailed comparison of our benchmark with existing benchmarks. 

\begin{table*}[t]
\centering
\setlength{\tabcolsep}{4pt}
\renewcommand{\arraystretch}{1.25}

\resizebox{\textwidth}{!}{%
\begin{tabular}{l*{12}{c}}
\toprule[1.2pt]
\makecell[l]{Scope /\\behavior}
& \makecell{Full-Duplex-\\Bench v1}
& \makecell{Full-Duplex-\\Bench v1.5}
& \makecell{Full-Duplex-\\Bench v2}
& \makecell{Full-Duplex-\\Bench v3}
& $\tau$-Voice
& EVA-Bench
& \makecell{VAmoS\\Bench}
& EchoChain
& IHBench
& \makecell{APEX-\\Voice}
& \makecell{VoiceAgent\\Bench}
& \textbf{\makecell{Our\\benchmark}} \\
\midrule

Evaluation scope
& CD & CD & CD & Both & Both & Both
& Both & CD & CD & Both & AP & Both \\

Interruption
& \yesmark & \yesmark & \yesmark & \nomark
& \yesmark & \yesmark & \yesmark$^{*}$
& \yesmark & \yesmark & \yesmark & \nomark & \yesmark \\

Resume
& \nomark & \nomark & \nomark & \nomark
& \nomark & \yesmark$^{**}$ & \nomark
& \yesmark & \yesmark & \nomark & \nomark & \yesmark \\

Priority change
& \nomark & \nomark & \nomark & \nomark
& \nomark & \nomark & \nomark
& \nomark & \nomark & \nomark & \nomark & \yesmark \\

Task addition
& \nomark & \nomark & \nomark & \nomark
& \yesmark & \nomark & \nomark
& \nomark & \nomark & \nomark & \nomark & \yesmark \\

Task withdraw
& \nomark & \nomark & \nomark & \nomark
& \nomark & \nomark & \nomark
& \nomark & \nomark & \yesmark & \nomark & \yesmark \\

Status inquiry
& \nomark & \nomark & \nomark & \nomark
& \nomark & \nomark & \nomark
& \nomark & \nomark & \nomark & \nomark & \yesmark \\

\bottomrule[1.2pt]
\end{tabular}%
}

\smallskip
\begin{minipage}{\textwidth}
\footnotesize
CD: conversational dynamics; AP: agentic performance;
Both: CD and AP.\\
$^{*}$Barge-in counts.
\quad $^{**}$Resumption latency only.
\end{minipage}
\caption{Comparison of evaluation scope and user behaviors across benchmarks.}
\label{tab:benchmark-comparison}
\end{table*}

\subsection{User Simulator}
Voice-agent benchmarks use prerecorded audio or user simulators to provide controlled inputs for scalable, reproducible evaluation. Some work \cite{lin2025fullv1,lin2026fullv15,lin2026fullv3,jain2025voiceagentbench,salimi2026ihbench} rely on fixed prerecorded inputs. However, these fixed audio inputs are limited in capturing the natural flow and contextual dependencies of multi-turn conversation \cite{lin2026fullv2}.Such limitations motivate interactive simulators that adapt to evolving user needs and agent behavior. 
Full-Duplex-Bench v2 \cite{lin2026fullv2} offers configurable pacing, while $\tau$-voice \cite{DBLP:journals/corr/abs-2603-13686} supports continuous full-duplex simulation with configurable turn-taking. EVA-Bench \cite{bogavelli2026eva} and VAmoS Bench \cite{meyer2026vamos} guide simulators with task goals and user profiles. EchoChain \cite{modi2026echochain} combines adaptive simulation with controlled interruption timing, while APEX-Voice \cite{mathur2026apex} selects prerecorded utterances by workflow state.  We extend the user simulator in $\tau$-voice \cite{DBLP:journals/corr/abs-2603-13686} to control intervention type and timing using tool execution state and actual audio playback progress. It also answers clarification requests, enabling targeted, traceable evaluation of coordination during ongoing work and interrupted result delivery.

\section{Method}
\begin{figure*}[t]
  \includegraphics[width=\linewidth]{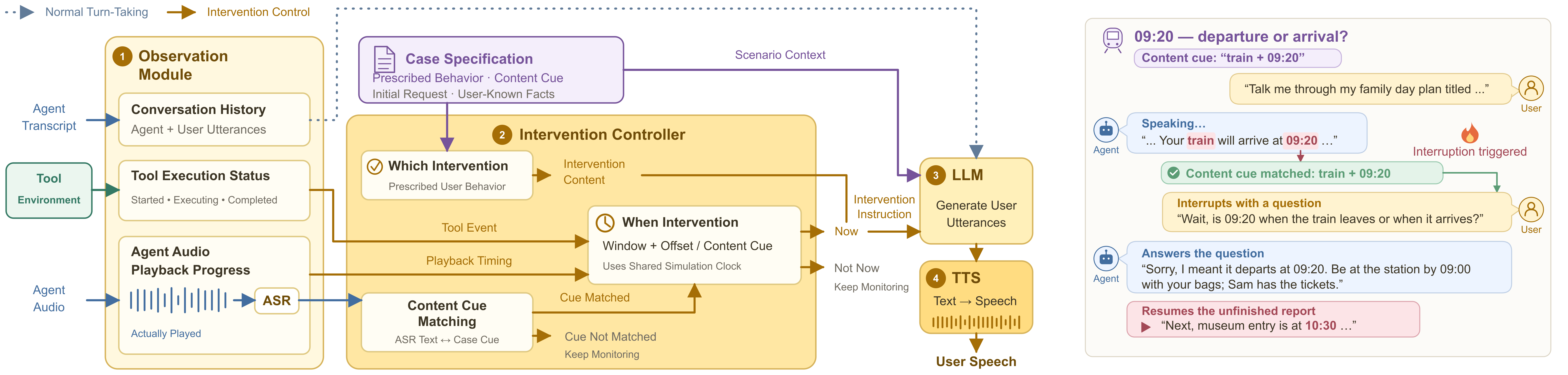} 
  \caption {User simulator architecture with a content-triggered inquiry–resume example.}
  \label{fig:user_simulator_architecture}
\end{figure*}

\subsection{Benchmark Framework}
\begin{table*}[t]
\centering
\small
\begin{tblr}{
  width = \linewidth,
  colspec = {X[1.2,l] X[1.7,l] X[3.2,l] X[4.2,l]},
  colsep = 4pt,
  rowsep = 5pt,
  row{1} = {font=\bfseries},
  cell{2}{1} = {r=5}{valign=m},
  cell{7}{1} = {r=2}{valign=m},
  hline{1,9} = {1.2pt},         
  hline{2,7} = {0.4pt},        
  hline{3-6,8} = {2-4}{0.4pt},  
}
Group
& Task
& Prescribed user behavior
& Successful agent behavior \\

Coordination during task execution
& Status Inquiry
& Issue one request then inquire about the execution status  twice during task execution.
& Answer the status inquiries and report the result. \\

& Baseline with no request changes
& Issue 5 requests with no further interventions.
& Execute tasks concurrently and report all required results. \\

& Priority change
& Issue 5 requests then ask for priority feedback on the result of a designated task during task execution.
& Prioritize that result in subsequent reporting and report the remaining results. \\

& Task withdraw
& Issue 5 requests then withdraw one task's request during task execution.
& Omit the withdrawn task and report the remaining results. \\

& Task addition
& Issue 5 requests then add one more task's request during task execution.
& Execute the added task and report all required results. \\

Coordination during spoken result delivery
& Pause-resume
& Interrupt the agent with a pause request during spoken result delivery.
& Pause appropriately and wait; once the user returns, resume and complete the result report. \\

& Inquiry-resume
& Interrupt the agent with a follow-up question during spoken result delivery.
& Answer the question, then resume and complete the result report. \\
\end{tblr}
\caption{Summary for the task conditions, prescribed user behaviors, and successful agent behaviors.}
\label{tab:task-design}
\end{table*}

Our benchmark evaluates whether full-duplex voice agents adapt to evolving requests while tracking task and communication progress. Tasks fall into two groups: (1) coordination during task execution, including a baseline without interventions, status inquiries, priority changes, request withdraw, and task addition; and (2) coordination during spoken result delivery, including pause-resume and inquiry-resume interactions during unfinished announcements. Table~\ref{tab:task-design} summarizes task conditions, prescribed user behaviors, and successful agent behaviors.

Figure \ref{fig:benchmark_framework} illustrates the benchmark framework, comprising a user simulator, an API-based voice agent, and an asynchronous tool environment. The simulator streams user speech to the agent, which produces spoken responses and tool calls. The environment executes these calls and returns their results while speech interaction continues. Recorded interactions support evaluation of execution responsiveness and coordination under request changes and interruptions.

\subsection{Controllable User Simulator}
\textbf{Tick-based Simulation Clock} 
Following $\tau$-Voice, we decouple simulated conversational time from wall-clock computation by pausing the simulation during non-realtime LLM response generation and TTS synthesis. Audio is exchanged in fixed-duration ticks (e.g., 200 ms), each advancing simulated time by that duration regardless of preparation latency. Both participants can speak within the same tick, enabling overlap and interruptions. This allows an LLM-TTS pipeline to simulate a full-duplex user with independent control over language generation, speech synthesis, and user behavior.
 
\noindent \textbf{User Simulator Architecture} As illustrated in Figure \ref{fig:user_simulator_architecture}, our user simulator comprises an observation module, an intervention controller, an LLM, and a TTS component. The observation module maintains the conversation history and tracks tool execution events and audio playback progress. Using these observations and the case specification, the intervention controller determines which prescribed user behavior to initiate and when. The LLM generates contextually appropriate utterances, which TTS converts into speech for buffering and streaming to the agent through the tick-based interaction loop.
 
\noindent \textbf{Intervention Timing Control} Controlling intervention timing is essential to achieving our coordination evaluation objectives. As shown in Figure~\ref{fig:user_simulator_architecture}, the intervention controller manages three timing aspects: (1) the tool execution window, identified from tool events, with interventions scheduled at configured offsets; (2) the spoken result delivery window, detected through incremental ASR of played audio after tool results become available; and (3) content-based interruption points within result delivery, triggered when ASR transcripts match specified content cues. Timing follows the shared simulation clock and reflects actual execution and playback progress.
 

\subsection{Evaluation Protocol}
\begin{table*}[t]
\centering
\footnotesize
\begin{tblr}{
  width = \linewidth,
  colspec = {X[1.3,l] X[3.4,l] X[5.3,l]},
  colsep = 4pt,
  rowsep = 4pt,
  row{1} = {font=\bfseries},
  cell{2}{1} = {r=5}{valign=m, font=\bfseries},
  cell{7}{1} = {r=2}{valign=m, font=\bfseries},
  cell{9}{1} = {r=3}{valign=m, font=\bfseries},
  hline{1,12} = {1.2pt},             
  hline{2,7,9} = {0.4pt},            
  hline{3-6,8,10-11} = {2-3}{0.4pt}, 
}
Dimension
& Metric name
& Definition \\

Execution responsiveness
& Tool start latency (s) $\downarrow$
& The time elapsed from the end of the user request or the completion of the latest relevant clarification associated with a tool, to when the agent invokes that tool. \\

& Announcement latency (s) $\downarrow$
& The time elapsed from successful tool completion to the start of the first substantive spoken announcement grounded in the returned tool result. \\

& Case--level first--attempt success rate (pass@1, \%) $\uparrow$
& Percentage of cases in which every expected tool is selected with correct arguments on its first attempt. Missing tools or incorrect first attempts cause the case to fail. \\

& Status Inquiry Response Rate (\%) $\uparrow$
& Percentage of cases in which the agent replies to the user's status check inquiry. \\

& Concurrent invocation rate (concurrency met, \%) $\uparrow$
& Percentage of cases in which the agent invokes tools concurrently when handling requests that require multiple tools. \\

Coordination under request changes
& Request change success rate (\%) $\uparrow$
& Percentage of cases in which the agent satisfies the changed request: announcing the prioritized result first after a priority change, cancelling the withdrawn result, or reporting the added result. \\

& Result completion rate (\%) $\uparrow$
& Percentage of cases in which all required tool results are successfully obtained and announced: five results for the baseline and priority change, four for task withdraw, and six for task addition. \\

Coordination under interruptions
& Interruption latency (s) $\downarrow$
& The time elapsed from the onset of the user's interruption to the point at which the agent is confirmed to have stopped announcing the result. \\

& Resumption rate (\%) $\uparrow$
& Percentage of cases in which the agent resumes after a pause interruption or follow-up question. Full restarts and incorrect claims that all required information has already been communicated count as failures. \\

& Information omission rate (\%) $\downarrow$
& (total number of expected items $-$ number of delivered items) $/$ total number of expected items

``Delivered items'' is the union of expected items completely delivered before the interruption and after resumption. Items spoken during the interruption do not count. \\
\end{tblr}
\caption{Definitions for the evaluation metrics.}
\label{tab:metric-design}
\end{table*}
\textbf{Voice Agent Evaluation} We evaluate agents' coordination of task and communication progress along three dimensions, with detailed metric definitions in Table~\ref{tab:metric-design}. Execution responsiveness measures how promptly agents invoke appropriate tools and report available results while other tools run. Coordination under request changes assesses adaptation to priority changes, task withdrawals, and task additions while preserving unaffected obligations. Coordination under interruptions assesses whether agents accurately track the interruption point and resume unfinished result delivery after a pause or follow-up question, distinguishing what the user has already heard from what remains to be communicated.
 
\noindent \textbf{User Simulator Controllability Evaluation} Since incorrect or mistimed user interventions can distort the evaluation of agent coordination, we access user-simulator controllability using three metrics. Behavior delivery rate measures the proportion of prescribed interventions delivered with the intended meaning. Intervention-window hit rate measures the proportion of prescribed interventions delivered within the designated time windows, using actual speech onset rather than trigger time. Clarification-response rate measures the proportion of clarification questions appropriately answered using case-provided information. These metrics verify whether the simulator realizes the intended evaluation conditions in both content and timing.

\section{Experimental Setup}
\subsection{Domains and Scenarios}
Our benchmark comprises 60 scenarios across 6 everyday domains: in-car assistance, cooking, retail, travel, telecommunications, and household management. The scenarios are designed to reflect everyday human needs and conversational behaviors. Each scenario groups related requests around a shared practical goal, such as preparing for a journey, organizing a meal, or managing family schedules. Request changes and interruptions have contextually plausible motivations: a user may prioritize charging availability when the battery is low, withdraw a check after a family member supplies the answer, or pause a report to attend to another activity. We instantiate these scenarios as 140 test cases. 20 multi-tool scenarios each yield a baseline with no request changes and three request-change variants-priority change, task withdraw, and task addition-producing 80 cases. 20 single-tool scenarios include user inquiries during execution, producing 20 cases. The remaining 20 result-delivery scenarios each yield pause-resume and inquiry-resume variants, producing 40 cases. Each interaction condition includes 20 cases, with variants of the same scenario sharing the initial request and tool configuration to support controlled comparisons as user requests change or result delivery is interrupted.

We assign controlled tool execution durations to represent operations with different response times and create opportunities for interaction while work is ongoing. In multi-tool scenarios, the five initial tools have execution durations of 60, 15, 5, 5, and 5 seconds, allowing some results to become available while other operations remain in progress. These durations are identical across the baseline with no request changes and the three request-change variants; the additional tool in task-addition cases takes 15 seconds. Single-tool waiting scenarios use a 60-second duration, providing a window for users to ask status during execution. For pause-resume and inquiry-resume scenarios, tools return five-item reports without an artificial execution delay, allowing evaluation to focus on coordination during spoken delivery. These controlled settings support consistent comparisons of how agents manage ongoing execution, available results, and evolving user requests.

\subsection{Configurations for API-based Voice Agent}
We evaluated five top-tier application-level close-source voice agents, all accessed through their official APIs. We evaluate these five voice agents in English using the same scenario definitions, tool interfaces, and reporting instructions. Each agent receives the complete set of 80 tools and is instructed to initiate independent tool calls concurrently, report available results promptly, and resume unfinished result delivery after interruptions. The evaluation includes 140 cases per agent, yielding 700 conversations, with a maximum conversation duration of 480 seconds.

Realtime 1.5, Realtime 2, Grok Voice 3.0, and Gemini 3.8 Live share a tick-based simulation clock, with each representing 0.2 seconds. All four use default VAD settings and interruption handling supported by their official API protocols. When an interruption is detected, the framework stops local playback and discards unplayed audio. Tool execution and result delivery follow each API's supported protocol. Realtime 2 uses high reasoning effort with parallel tool calling enabled, while Realtime 1.5 retains its default tool-calling configuration. Both receive tool results as they become available, with explicit requests to continue responding. Gemini uses non-blocking function calls and WHEN\_IDLE result scheduling, allowing results to arrive independently while scheduling their announcement after the current response. Grok also executes tool calls asynchronously, but its protocol requires all outstanding function results to be submitted before requesting a spoken continuation. We additionally wait for ongoing playback to finish because preliminary attempts to announce individual results immediately produced overlapping speech. GPT Live 1.0 uses its native continuous audio interface, paired with GPT-5.6-Terra as the reasoning and tool-use backend, with parallel tool calling enabled and no separately configured VAD component. It requires a continuous input audio stream paced in real time, preventing the use of the same pausable tick-based simulation clock as the other four APIs. 

\subsection{Configurations for User Simulator}
The user simulator uses GLM-5.3 with temperature 1.0 to generate contextually appropriate responses, including answers to clarification questions, and Qwen3-TTS-12Hz-1.7B-Base to synthesize English speech with a shared reference voice. Dialogue context is constructed from playback-aligned assistant transcripts. For inquiry-triggered interruptions, Qwen3-ASR-1.7B transcribes played assistant audio, and GPT-5.6-Terra determines whether the designated result content has been reached. 

\subsection{Configurations for Auto-evaluation}
For automated evaluation, we transcribe recorded client-side playback using Qwen3-ASR-1.7B and obtain word-level timestamps with Qwen3-ForcedAligner-0.6B, using Whisper-large-v3 as a fallback when transcription or timestamp validation fails. The transcripts are combined with tool-call records, returned results, and interruption events to assess tool-call correctness, compliance with changed requests, resumption, and information coverage. Claude Opus 5 first serves as the semantic evaluator, identifying relevant utterances and matching announced information against the required tool results. Latencies and aggregate scores are then computed programmatically from recorded timestamps and validated annotations. GPT 6 Astra subsequently reviews the annotations and evaluation outputs for correctness, with any necessary corrections incorporated into the final reported results.

\section{Results and Analysis}
\subsection{Overall Performance}

\begin{table*}[t]
\centering
\footnotesize
\begin{tblr}{
  width = \linewidth,
  colspec = {
    X[1.45,l,m]
    X[1.10,c,m]
    X[1.20,c,m]
    X[1.00,c,m]
    X[1.20,c,m]
    X[0.90,c,m]
    X[0.90,c,m]
    X[0.90,c,m]
    X[1.00,c,m]
  },
  colsep = 2pt,
  rowsep = 2pt,
  row{1,2} = {font=\scriptsize\bfseries},
  cell{1}{2} = {c=3}{c,m},
  cell{1}{5} = {c=2}{c,m},
  cell{1}{7} = {c=3}{c,m},
  hline{1,8} = {1pt},
  hline{2,3} = {0.4pt},
  vline{2,5,7} = {0.4pt},
}
& {Execution responsiveness}
& &
& {Coordination under request changes}
&
& {Coordination under interruptions}
& & \\

Model
& {Tool start\\latency S/M $\downarrow$}
& {Announcement\\latency S/M $\downarrow$}
& {Status inquiry\\response $\uparrow$}
& {Request change\\success $\uparrow$}
& {Result\\completion $\uparrow$}
& {Information\\omission $\downarrow$}
& {Resumption $\uparrow$}
& {Interruption\\latency $\downarrow$} \\

GPT Live 1.0
& 3.33 / 6.64s
& \textbf{3.89} / 33.74s
& 92.5\%
& 15.0\%
& 30.0\%
& \textbf{3.5\%}
& \textbf{90.0\%}
& 2.53s \\

Realtime 1.5
& 1.88 / 37.37s
& 6.70 / 32.73s
& \textbf{100\%}
& \textbf{85.0\%}
& \textbf{55.0\%}
& 34.5\%
& 20.0\%
& 1.15s \\

Realtime 2
& 3.91 / 5.37s
& 7.33 / \textbf{17.59}s
& 97.37\%
& 10.0\%
& 40.0\%
& 79.0\%
& 0.0\%
& \textbf{1.07}s \\

Grok Voice 3.0
& 2.86 / \textbf{4.73}s
& 6.07 / 51.16s
& \textbf{100\%}
& 30.0\%
& 30.0\%
& 23.0\%
& 55.0\%
& 1.69s \\

Gemini 3.8 Live
& \textbf{1.19} / 5.62s
& 4.05 / 22.96s
& \textbf{100\%}
& 70.0\%
& 15.0\%
& 54.5\%
& 60.0\%
& 2.38s \\
\end{tblr}
\caption{Overall performance of 5 agents on our benchmark. S/M denotes the results for single-tool/multi-tool tasks.}
\label{tab:overall-performance}
\end{table*}

Table \ref{tab:overall-performance} summarizes the overall performance. These results reveal limitations in current voice agents’ ability to track and coordinate ongoing execution, completed work, communicated information, and outstanding obligations as the conversation evolves. For execution responsiveness, all five agents exhibit higher tool start latency and result announcement latency on multi-tool tasks than on single-tool tasks. This suggests that handling more tasks still places a greater burden on voice agents when interpreting user requests, tracking execution status, and coordinating result announcements. Request changes further challenge agents’ ability to adapt to new user requests while continuing to address previously requested tasks. Three of the five agents achieve request change success rates of at most 30\%, while even the highest result completion rate is only 55\% across the three request change types. Moreover, Gemini 3.8 Live achieves a request change success rate of 70\% but a result completion rate of only 15\%, demonstrating that following changed requests does not ensure that all outstanding reporting obligations are fulfilled. Coordination under interruptions presents two challenges: resuming unfinished result delivery and accurately distinguishing what has already been communicated from what remains to be reported. GPT Live 1.0 achieves a resumption rate of 90\% and an information omission rate of 3.5\%, whereas the other four agents have resumption rates of at most 60\% and information omission rates of at least 23\%. These results indicate substantial difficulties in retaining unfinished reporting obligations and delivering information that the user has not yet heard. 

\subsection{Multi-Tool Scheduling and Coordination}
\begin{table}[t]
\centering
\footnotesize
\begin{tblr}{
  colspec = {l c c},
  colsep = 2pt,
  rowsep = 2pt,
  row{1} = {font=\scriptsize\bfseries},
  hline{1,7} = {1.2pt},
  hline{2} = {0.4pt},
}
Model
& Concurrency met $\uparrow$
& Pass@1 S/M $\uparrow$ \\

GPT Live 1.0
& 100.0\%
& 100.00\% / 88.75\% \\

Realtime 1.5
& 0.0\%
& 100.00\% / 63.75\% \\

Realtime 2
& 100.0\%
& 98.33\% / 86.25\% \\

Grok Voice 3.0
& 100.0\%
& 96.67\% / 76.25\% \\

Gemini 3.8 Live
& 100.0\%
& 96.67\% / 81.25\% \\
\end{tblr}
\caption{Performance comparison for single-tool and multi-tool tasks.}
\label{tab:multi_vs_single}
\end{table}

Table \ref{tab:multi_vs_single} with tool-start latency and announcement latency results in Table \ref{tab:overall-performance} reveals limitations in scheduling both tool invocation and result delivery. Realtime 1.5 invokes tools sequentially, waiting for the preceding tool result before initiating another tool call. This sequential scheduling results in a concurrency met score of 0\%, compared with 100\% for the other four voice agents, and contributes to its high multi-tool start latency of 37.37 s. Grok Voice 3.0 supports concurrent tool invocation, but following the provider's documented protocol, it should be configured to wait for all tool results to arrive and any ongoing speech to finish before announcing the results. We also tested announcing each result as it became available, but this caused overlapping announcements that severely disrupted result delivery. This result delivery waiting contributes to Grok's highest multi-tool result announcement latency of 51.16 s.

Across all five systems, multi-tool workloads yield lower Pass@1 and higher tool start latency and result announcement latency than single-tool tasks. These results suggest that multi-tool tasks impose greater demands on user intent understanding and coordination between tool invocation and result delivery. Our qualitative review of the recordings reveals more frequent requests for clarification of tool arguments in multi-tool tasks, including requests for users to explicitly spell out parameter values. This behavior suggests that agents have greater difficulty resolving the required tool arguments when handling multiple tools, consistent with the lower first-attempt accuracy. Clarification exchanges can also increase result-announcement latency: when one tool returns its result, the agent may still be clarifying arguments for another tool, delaying the announcement of the available result.

\subsection{Coordination Under Request Changes}
\begin{table*}[t]
\centering
\footnotesize
\begin{tblr}{
  colspec = {
    Q[l,m]
    Q[c,m]
    Q[c,m]
    Q[c,m]
    Q[c,m]
    Q[c,m]
    Q[c,m]
    Q[c,m]
  },
  colsep = 3pt,
  rowsep = 3pt,
  row{1,2} = {font=\scriptsize\bfseries},
  cell{1}{2} = {c=4}{c,m},
  cell{1}{6} = {c=3}{c,m},
  hline{1,8} = {1.2pt}, 
  hline{2,3} = {0.4pt}, 
  vline{2,6} = {0.4pt}, 
}
& Result completion rate $\uparrow$
& & &
& Request change success rate $\uparrow$
& & \\

Model
& Baseline
& Priority change
& Task withdraw
& Task addition
& Priority change
& Task withdraw
& Task addition \\

GPT Live 1.0
& 90.0\% & 85.0\% & 90.0\% & 55.0\%
& 80.0\% & 40.0\% & 75.0\% \\

Realtime 1.5
& 85.0\% & 80.0\% & 80.0\% & 75.0\%
& 95.0\% & 90.0\% & 100.0\% \\

Realtime 2
& 85.0\% & 75.0\% & 85.0\% & 65.0\%
& 10.0\% & 95.0\% & 95.0\% \\

Grok Voice 3.0
& 80.0\% & 60.0\% & 65.0\% & 50.0\%
& 35.0\% & 85.0\% & 100.0\% \\

Gemini 3.8 Live
& 60.0\% & 40.0\% & 50.0\% & 50.0\%
& 75.0\% & 90.0\% & 100.0\% \\
\end{tblr}
\caption{Performance for coordination under request changes.}
\label{tab:request-change-performance}
\end{table*}

Table \ref{tab:request-change-performance} evaluates how voice agents coordinate result delivery when users change their requests during tool execution. Request change success rate measures whether agents adapt result delivery to changed requirements: announcing the prioritized task's result first, suppressing the withdrawn task's result, or reporting the added task's result. Result completion rate measures whether all required results are announced, assessing whether agents preserve earlier reporting obligations as requests evolve. For every agent, result completion rates under all three request change conditions are no higher than the baseline without request changes. Averaged across agents, the rate falls from 80\% in the baseline to 68\% under priority change, 74\% under task withdraw, and 59\% under task addition. These results suggest that adapting to changing requirements introduces additional challenges for complete result delivery. 

Realtime 1.5 exhibits the most balanced performance, achieving request change success rates of 90–100\% and result completion rates of 75–80\% across the three request change types. The other four agents exhibit distinct weaknesses. GPT Live achieves a result completion rate of 90\% under task withdraw but a withdraw success rate of only 40\%, indicating that it frequently continues reporting results the user no longer wants. Realtime 2 and Grok struggle to report the prioritized result first, achieving priority change success rates of only 10\% and 35\%, respectively. Gemini follows revised requests relatively well, with request change success rates of 75–100\%, but its result completion rates remain at 40–50\%, revealing difficulty fulfilling all outstanding reporting obligations. Our qualitative review of the recordings further identifies a recurring failure pattern in Grok and Gemini: after a request change, they hallucinate results for other tools, sometimes announcing hallucinated tool result before the corresponding tools have finished executing. This behavior suggests difficulty distinguishing completed work from execution still in progress. Task addition exposes another coordination failure: all five agents achieve addition success rates of 75–100\%, yet complete all required results in only 50–75\% of cases. Realtime 1.5, Grok, and Gemini report the added result in every case, but their result completion rates are only 75\%, 50\%, and 50\%, respectively. Thus, successfully addressing a new request does not ensure that earlier requests are fully addressed.

\subsection{Coordination Under Interruptions}
\begin{table}[t]
\centering
\footnotesize
\begin{tblr}{
  width = \columnwidth,
  colspec = {
    Q[l,m]
    X[1.0,c,m]
    X[1.2,c,m]
  },
  colsep = 1pt,
  rowsep = 3pt,
  row{1} = {font=\scriptsize\bfseries},
  row{2} = {font=\scriptsize},
  cell{1}{1} = {r=2}{l,m},
  hline{1,8} = {1.2pt},
  hline{3} = {0.4pt},
}
Model
& Resumption rate $\uparrow$
& \mbox{Information omission rate $\downarrow$} \\

& Pause- / Inquiry-resume
& Pause- / Inquiry-resume \\

GPT Live 1.0
& 100.0\% / 90.0\%
& 0.0\% / 7.0\% \\

Realtime 1.5
& 80.0\% / 30.0\%
& 8.0\% / 61.0\% \\

Realtime 2
& 5.0\% / 0.0\%
& 78.0\% / 80.0\% \\

Grok Voice 3.0
& 95.0\% / 55.0\%
& 6.0\% / 40.0\% \\

Gemini 3.8 Live
& 90.0\% / 70.0\%
& 43.0\% / 66.0\% \\
\end{tblr}
\caption{Performance under pause-resume and inquiry-resume interruptions.}
\label{tab:interruption-performance}
\end{table}

Table \ref{tab:interruption-performance} evaluates two complementary aspects of coordination after an interruption. Resumption rate measures whether agents return to unfinished result delivery. Information omission rate assesses whether agents accurately track where result delivery was interrupted, distinguishing what had already been communicated from what still needed to be reported to the user. Although the agent’s system prompt explicitly requires it to resume execution in both pause–resume and inquiry–resume tasks, all five agents achieve lower resumption rates after an inquiry. Averaged across agents, the rate falls from 74\% under pause-resume to 49\% under inquiry-resume. A possible explanation is that a pause more clearly signals the intention to continue the interrupted delivery, whereas an inquiry introduces another conversational objective that the agent must address while retaining its earlier reporting obligation. Indeed, every agent successfully answers the intervening inquiry, but this does not ensure a return to unfinished delivery.

All five agents exhibit varying degrees of difficulty accurately tracking the interruption point and distinguishing what has already been communicated from what remains to be reported. To track this point accurately, agents must notice the client’s actual playback progress and the user’s accurate barge-in position within the result announcement. Audio transmitted to the client may not yet have been played, so treating all transmitted audio as already communicated can cause agents to skip information the user has not heard. GPT Live 1.0 resumes result delivery most effectively, with information omission rates of 0\% under pause-resume and 7\% under inquiry-resume. Realtime 2 performs worst on both metrics, with resumption rates of only 5\% and 0\% and information omission rates of 78\% and 80\% under pause-resume and inquiry-resume, respectively. Moreover, resuming speech does not ensure that all required information is announced: Gemini achieves a resumption rate of 90\% after a pause but an information omission rate of 43\%. These findings highlight two challenges for voice agent design: reliably resuming interrupted result delivery and ensuring that all remaining information is communicated to the user.

\subsection{User Simulator Controllability}
We assess whether the user simulator can reliably deliver prescribed behaviors at designated stages of the interaction while responding to clarification requests. Across 700 conversations, the simulator achieves a behavior delivery rate and an intervention-window hit rate of 99.86\% (699/700). In the single problematic case, the agent made incorrect tool calls and repeatedly requested clarification of a parameter until the time limit was reached, preventing complete delivery of the scheduled user behaviors. Among the 69 conversations in which agents asked clarification questions, the simulator successfully answered all 92 questions, yielding a clarification-response rate of 100\%. These results demonstrate reliable control over user behavior and intervention timing in the evaluated conversations, while retaining the flexibility to answer agents' questions as the interaction unfolds.

\section{Conclusion}
We introduced Joi-Bench-Task, a benchmark for evaluating how full-duplex voice agents coordinate ongoing task execution and spoken result delivery as user requests evolve. Our proposed user simulator reliably controls intervention content and timing while remaining interactive during conversations, establishing controllable and reproducible conditions for evaluating voice-agent coordination. Evaluating five leading commercial voice agents reveals persistent coordination failures. Multi-tool experiments indicate that handling multiple tasks places a greater burden on agents’ request interpretation and coordination of tool execution and result delivery. Experiments under request changes reveal difficulties adapting to revised requirements while preserving unaffected obligations. Interruption experiments expose failures to resume unfinished result delivery and communicate complete information. These findings point toward improving how voice agents track and coordinate ongoing execution, completed work, information communicated, and information remains to be reported. Joi-Bench-Task provides a framework for evaluating this coordination throughout evolving conversations.

\section*{Limitations}
Our benchmark currently focuses on English interactions. Extending the evaluation to additional languages would help assess whether the observed coordination patterns hold across different linguistic settings, including variations in how users express requests, seek clarification, and interrupt ongoing speech. Future work will explore multilingual evaluation to broaden the coverage of user interactions.

Our evaluation also uses audio without added background noise or environmental sounds, providing controlled acoustic conditions for studying task coordination. Everyday interactions may involve traffic noise, household sounds, or background conversations, which could influence request understanding and interruption handling. We plan to incorporate realistic environmental audio tailored to different usage scenarios, such as in-car assistance and cooking, to better represent the settings in which users interact with voice agents.

Finally, the tool environment uses mock functions with manually specified execution durations. This design supports reproducible comparisons and controlled interventions during ongoing execution. Integrating real tools and databases would complement this setup by introducing naturally varying response times and data-dependent results. Future work will explore these integrations to assess coordination in practical workflows while retaining the benchmark's controllability and reproducibility.

\bibliography{custom}

\appendix

\end{document}